\documentclass[final]{cvpr}

\usepackage{times}
\usepackage{epsfig}
\usepackage{graphicx}
\usepackage{amsmath}
\usepackage{amssymb}
\usepackage{mathrsfs}
\usepackage{booktabs}
\usepackage{tabularx}
\usepackage[para,online,flushleft]{threeparttable}
\usepackage{multirow}
\usepackage{caption}
\usepackage{subcaption}
\usepackage[ruled,linesnumbered]{algorithm2e}
\usepackage{color, xcolor}
\usepackage[title]{appendix}
\DeclareMathOperator*{\argmin}{arg\,min}

\usepackage[pagebackref=true,breaklinks=true,colorlinks,bookmarks=false]{hyperref}

\definecolor{mypink}{RGB}{219, 48, 122}
 
\colorlet{dark-blue}{blue!70!black}
\hypersetup{
    colorlinks=true,%
    citecolor=dark-blue,%
    filecolor=dark-blue,%
    linkcolor=red,%
    urlcolor=magenta
}

\def\cvprPaperID{837} 
\def\confYear{CVPR 2021}

\begin{document}
\title{Learning to Generalize Unseen Domains via Memory-based Multi-Source Meta-Learning for Person Re-Identification}
	
\author{Yuyang Zhao$^{\textcolor{mypink}{1}}$\thanks{Equal contribution. This work was done when Yuyang Zhao (yuyangzhao98@gmail.com) was a visiting student at Xiamen University.}~, 
Zhun Zhong$^{\textcolor{mypink}{2}*}$, 
Fengxiang Yang$^{\textcolor{mypink}{1}}$,
Zhiming Luo$^{\textcolor{mypink}{1}}\thanks{Corresponding author: \{zhiming.luo, szlig\}@xmu.edu.cn}$~, Yaojin Lin$^{\textcolor{mypink}{4}}$, Shaozi Li$^{\textcolor{mypink}{1,3}\dag}$, Nicu Sebe$^{\textcolor{mypink}{2}}$ \\
 \small{\textcolor{mypink}{1} Department of Artificial Intelligence, School of Informatics, Xiamen University} \\
 \small{\textcolor{mypink}{2} Department of Information Engineering and Computer Science, University of Trento}\\ 
 \small{\textcolor{mypink}{3} Institute of Artificial Intelligence, Xiamen University} \\
 \small{\textcolor{mypink}{4} Minnan Normal University} \\
\small{Project: \url{https://github.com/HeliosZhao/M3L}}
}

\maketitle
	
	\begin{abstract}
		
		Recent advances in person re-identification (ReID) obtain impressive accuracy in the supervised and unsupervised learning settings. 
		However, most of the existing methods need to train a new model for a new domain by accessing data. 
		Due to public privacy, the new domain data are not always accessible, leading to a limited applicability of these methods. In this paper, we study the problem of multi-source domain generalization in ReID, which aims to learn a model that can perform well on unseen domains with only several labeled source domains. 
		To address this problem, we propose the \textbf{M}emory-based \textbf{M}ulti-Source \textbf{M}eta-\textbf{L}earning (M$^3$L) framework to train a generalizable model for unseen domains. Specifically, a meta-learning strategy is introduced to simulate the train-test process of domain generalization for learning more generalizable models. To overcome the unstable meta-optimization caused by the parametric classifier, we propose a memory-based identification loss that is non-parametric and harmonizes with meta-learning. We also present a meta batch normalization layer (MetaBN) to diversify meta-test features, further establishing the advantage of meta-learning. Experiments demonstrate that our M$^3$L can effectively enhance the generalization ability of the model for unseen domains and can outperform the state-of-the-art methods on four large-scale ReID datasets. 
		
	\end{abstract}
	
	\section{Introduction}
	
	Person re-identification (ReID) aims at matching persons of the same identity across multiple camera views. Recent works in ReID mainly focus on three settings, \ie, fully-supervised~\cite{zhang2020relation,zheng2019joint,zhou2019omni}, fully-unsupervised~\cite{lin2019bottom, lin2020unsupervised,wang2020unsupervised} and unsupervised domain adaptive~\cite{fu2019self,zhai2020multiple,zhong2019invariance} ReID. Despite their good performance on a seen domain (\ie, a domain with training data), most of them suffer from drastic performance decline on unseen domains. In real-world applications, the ReID systems will inevitably search persons in new scenes. Therefore, it is necessary to learn a model that has good generalization ability to unseen domains.
	
	\begin{figure}[t]
	    \centering
	    \begin{subfigure}[t]{0.45\textwidth}
            \centering
            \includegraphics[width=1\textwidth]{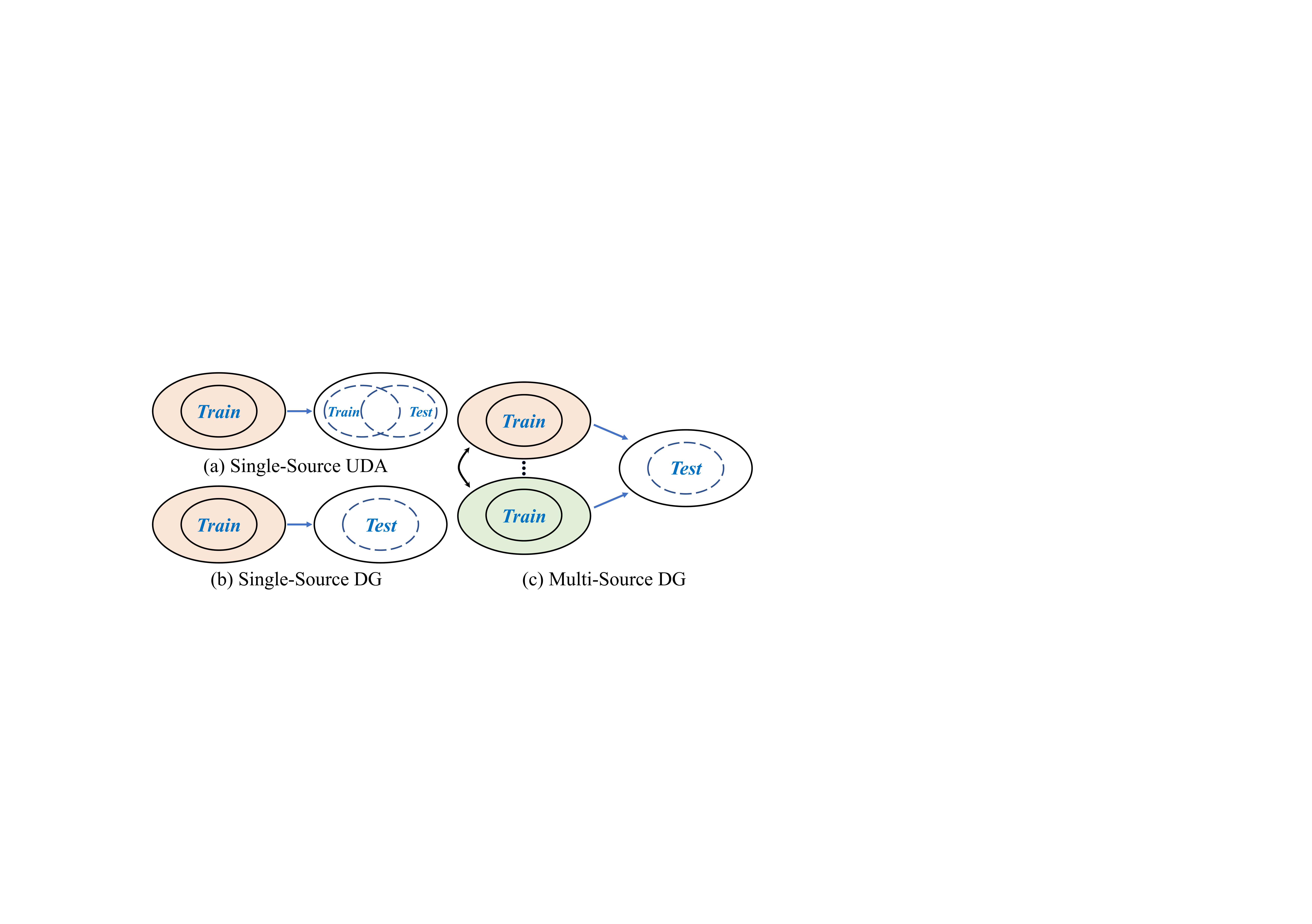}
            \subcaption*{}
        \end{subfigure}
        \quad
	    \begin{subfigure}[t]{0.45\textwidth}
	    \begin{center}
	    \fontsize{8pt}{9pt}\selectfont
	    \vspace{-.2in}
	    \subcaption*{Comparison of different settings in ReID.}
	    \vspace{-.1in}
    	\begin{tabular}{p{2.5cm}|p{0.4cm}<{\centering}p{0.5cm}<{\centering}p{0.5cm}<{\centering}p{0.5cm}<{\centering}p{0.5cm}<{\centering}} 
    		\toprule
    		\multirow{2}{*}{Setting} & \multicolumn{3}{c}{Source(s)} & \multicolumn{2}{c}{Target(s)} \\
    		~ & Multi & Images & Labels & Images & Labels  \\
    		\midrule
    		Fully-Supervised &$\times$ & \checkmark & \checkmark & --- & --- \\
    		Fully-Unsupervised 	& $\times$ &\checkmark & $\times$ & --- & --- \\
    		Single-Source UDA & $\times$ & \checkmark & \checkmark & \checkmark & $\times$ \\
    		Single-Source DG & $\times$ & \checkmark & \checkmark & $\times$ & $\times$ \\
    		Multi-Source DG & \checkmark & \checkmark & \checkmark & $\times$ & $\times$ \\
    		\bottomrule
    	\end{tabular} 
	    \end{center}
	    \end{subfigure}
	    \vspace{-.1in}
	    \caption{Comparison of different settings in person ReID. Different background colors indicate different distributions, \ie, domains. Solid/dashed ellipses denote data subset with/without labels. Domain generalization (DG) is designed to learn models for unseen domains, while other settings focus on learning models for specific domains. Compared to single-source DG, multi-source DG leverages knowledge from multiple labeled datasets, enforcing the model to learn more underlying patterns across domains.}
	    \label{fig:DG}
	\end{figure}
	
	To meet this goal, domain generalization (DG) is a promising solution that aims to learn generalizable models with one or several labeled source domains. As shown in Fig.~\ref{fig:DG}, compared to other settings, DG does not require the access to target domains. Generally, DG can be divided into two categories, single-source DG ~\cite{jin2020style,Liao2020QAConv,zhou2019learning} and multi-source DG~\cite{kumar2019fairest,song2019generalizable}, according to the number of source domains.
	Recent works mainly focus on single-source DG where only one labeled source domain is available. However, a single domain provides limited training samples and scene information, restricting the improvement of single-source DG methods. In contrast, multi-source DG utilizes multiple datasets of different distributions, providing more training data that contain numerous variations and environmental factors. However, due to the strong compatibility of deep networks, directly aggregating all source domains together might lead the model to overfit on the domain bias, hampering the generalization ability of the model. Although we can sample balanced training data from all source domains during training to reduce the impact of domain bias, the above issue still remains.
	
	In this paper, we study the multi-source DG and aim to enforce the model to learn discriminative features without domain bias so that the model can be generalized to unseen domains. 
	To achieve this goal, this paper introduces a meta-learning strategy for multi-source DG, which simulates the train-test process of DG during model optimization. In our method, we dynamically divide the source domains into meta-train and meta-test sets at each iteration. The meta-train is regarded as source data, and the meta-test is regarded as ``unseen'' data. During training, we encourage the loss of meta-train samples to optimize the model towards a direction that can simultaneously improve the accuracy of meta-test samples. 
	Nevertheless, meta-learning causes a problem for traditional parametric-based identification loss --- unstable optimization. On the one hand, ReID datasets contain numerous IDs, so the number of classifier parameters will surge when multiple domains are used for training. 
	On the other hand, the unified optimization of classifiers is unstable due to the asynchronous update by the high-order gradients of the meta-test. 
	Consequently, we propose a memory-based identification loss, which uses a non-parametric memory to take full advantage of meta-learning while avoiding unstable optimization. 
	We also introduce a meta batch normalization layer (MetaBN), which mixes meta-train knowledge with meta-test features to simulate the feature variations in different domains.
	Our full method is called \textbf{M}emory-based \textbf{M}ulti-Source \textbf{M}eta-\textbf{L}earning (M$^3$L). Experiments on four large-scale ReID datasets demonstrate the effectiveness of our M$^3$L when testing on unseen domains and show that our M$^3$L can achieve state-of-the-art results.
	
	Our contributions are summarized as follows:
	\begin{itemize}
	
		\item We propose a Multi-Source Meta-Learning framework for multi-source DG, which can simulate the train-test process of DG during training. Our method enables the model to learn domain-invariant representations and thus improves the generalization ability.
		
		\item We equip our framework with a memory-based module, which implements the identification loss in a non-parametric way and can prevent unstable optimization caused by traditional parametric manner during meta-optimization.
		
		\item We present MetaBN to generate diverse meta-test features, which can be directly injected into our meta-learning framework and obtain further improvement.
		
	\end{itemize}

	\section{Related Work}
	\label{sec:relatedwork}
	\textbf{Person Re-identification.} Recently, supervised learning approaches~\cite{chen2019abd,suh2018part,tay2019aanet,wang2018learning,zhang2020relation,zheng2019joint,zhong2020random,zhong2019camstyle,zhou2019omni} have achieved significant performance in person re-identification (ReID), relying on labeled training data. Considering the difficulties and complexities of annotations, unsupervised learning (USL)~\cite{fan2018unsupervised,lin2019bottom,lin2020unsupervised,wang2020unsupervised} and unsupervised domain adaptation (UDA)~\cite{chen2019instance, fu2019self,zhai2020ad,zhai2020multiple,zhong2018generalizing,zhong2019invariance,zou2020joint} methods are proposed. UDA aims to utilize labeled source data and unlabeled target data to improve the model performance on the target domain. UDA methods mainly focus on generating pseudo-labels on target data ~\cite{fu2019self, zhai2020multiple,zhong2019invariance} or transferring source images to the styles of the target domain for providing extra supervision during adaptation ~\cite{chen2019instance,zhong2018generalizing,zou2020joint}.
    USL approaches learn discriminative features only from unlabeled target data, the mainstream~\cite{fan2018unsupervised,lin2019bottom} of which is to train models with pseudo-labels obtained by clustering.
	\par
	\textbf{Domain Generalization.} Although USL and UDA ReID methods show good performance, they still need to collect a large amount of target data for training models. In contrast, domain generalization (DG) has no access to any target domain data. By carefully designing, DG methods~\cite{2020EccvDMG,jin2020style,Li2018MLDG} can improve the model performance on unseen domains. Most existing DG methods focus on closed-set tasks~\cite{2020EccvDMG,khosla2012undoing,Li2018MLDG,muandet2013domain,qiao2020learning}, assuming that the target data have the same label space as the source data. 
	Lately, several works~\cite{jin2020style,Liao2020QAConv,song2019generalizable,zhou2019learning} were introduced to learn generalizable models for person ReID.
	SNR~\cite{jin2020style} disentangles identity-relevant and identity-irrelevant features and reconstructs more generalizable features. 
	Liao \etal~\cite{Liao2020QAConv} propose a novel QAConv for calculating the similarity between samples, which can effectively improve ReID accuracy in unseen data but is inefficient during testing.
	DIMN~\cite{song2019generalizable} proposes a mapping subnet to match persons within a mini-batch and trains the model with data from one domain at each iteration. Song \etal~\cite{song2019generalizable} claim that DIMN uses meta-learning in the training stage. However, DIMN optimizes the model with the common training strategy, which is completely different from our meta-learning strategy.
	\par
	\begin{figure*}[t]
		\begin{center}
			\includegraphics[width=0.92\linewidth]{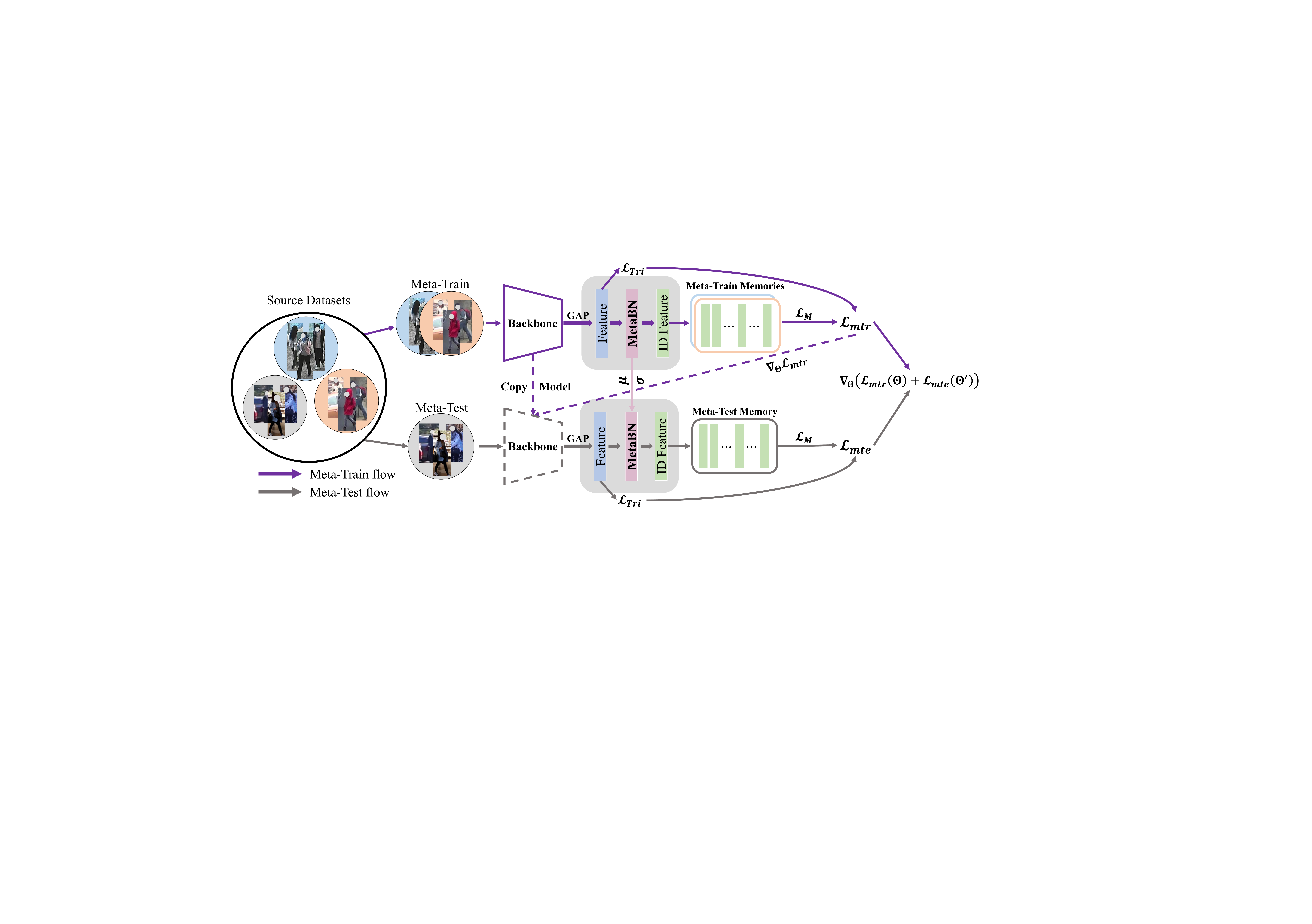}
		\end{center}
		\vspace{-7mm}
		\caption{The framework of the proposed M$^3$L. During training, we are given several (three in this example) source domains. At each iteration, source domains are divided into one meta-test and two meta-train domains. In the meta-train stage, memory-based identification loss and triplet loss are calculated from meta-train data as the meta-train loss. In the meta-test stage, the original model is copied and then the copied model is updated with meta-train loss. We compute the meta-test loss on the updated model. In this stage, MetaBN is used to diversify the meta-test features. Finally, the combination of meta-train and meta-test losses is used to optimize the original model.}
		\label{fig:overall}
		\vspace{-1mm}
	\end{figure*}
	
	\textbf{Meta Learning.} The concept of meta-learning~\cite{thrun1998learning} is learning to learn, and has been initially proposed in the machine learning community. Recently, meta-learning has been applied to various deep-based applications, including model optimization~\cite{andrychowicz2016learning,li2016learning}, few-shot learning~\cite{finn2017model,snell2017prototypical,sun2019meta,vinyals2016matching} and domain generalization~\cite{balaji2018metareg,guo2020learning,Li2018MLDG,Li_2019_ICCV,li2019feature}. 
	MAML~\cite{finn2017model} and its variant Reptile~\cite{nichol2018first} are proposed to learn a good initialization for fast adapting a model to a new task. 
	Li \etal~\cite{Li2018MLDG} first extend MAML~\cite{finn2017model} to closed-set DG. 
	Latter, meta-learning was applied to closed-set DG~\cite{balaji2018metareg,Li_2019_ICCV,li2019feature} and open-set DG~\cite{guo2020learning}. In this paper, we propose a memory-based meta-learning approach, which is tailor-made for multi-source DG in ReID.
 
	\section{Methodology}
	
	For multi-source domain generalization (DG) in person ReID, we are provided with $N_S$ source domains $\mathcal D_S = \{\mathcal D^1_S, ...,\mathcal D^{N_S}_S\}$ in the training stage. The label spaces of the source domains are disjointed. The goal is to train a generalizable model with the source data. In the testing stage, the model is evaluated directly on a given unseen domain $\mathcal D_T$.

	\subsection{Overview}
	  This paper designs a Memory-based Multi-source Meta-Learning (M$^3$L) framework for multi-source domain generalization (DG) in person ReID task.
	  In our framework, we introduce a meta-learning strategy, which simulates the train-test process of DG during model optimization. 
	  Specifically, we dynamically split the source domains into meta-train and meta-test at each iteration. During training, 
	  we first copy the original model and update it with the loss from meta-train data. Then we use the updated model to compute the meta-test loss.
	  The memory-based identification loss and triplet loss are adopted for effective meta-learning. We also inject a meta batch normalization layer (MetaBN) into the network, which diversifies the meta-test features with meta-train distributions to further facilitate the effect of meta-learning. 
	  Finally, the combination of the meta-train and meta-test losses is used to update the original model towards a generalizable direction that performs well on meta-train and meta-test domains.

    \subsection{Meta-Learning for Multi-Source DG}
	\label{sec:metaoptimization}
 
	We adopt the concept of ``learning to learn'' to simulate the train-test process of domain generalization during the model optimization. At each training iteration, we randomly divide $N_S$ source domains into $N_S-1$ domains as \textbf{meta-train} and the rest \emph{one} domain as \textbf{meta-test}. The process of computing the meta-learning loss includes the meta-train and the meta-test stages.

	\textit{In the meta-train stage}, we calculate the meta-train loss $\mathcal{L}_{mtr}$ on the meta-train samples to optimize the model. \textit{In the meta-test stage}, the optimized model is used to calculate the meta-test loss $\mathcal{L}_{mte}$ with the meta-test samples.
	Finally, the network is optimized by the combination of meta-train and meta-test losses, \ie,
	\begin{equation}
	\label{loss:meta}
	\argmin_{\Theta} \mathcal{L}_{mtr}(\Theta) + \mathcal{L}_{mte}(\Theta^{'}),
	\end{equation}
	where $\Theta$ denotes the parameters of the network, and $\Theta^{'}$ denotes the parameters of the model optimized by the $\mathcal{L}_{mtr}$. 
	Note that, $\mathcal{L}_{mte}$ is only used to update $\Theta$, the derivative of which is the high-order gradients on $\Theta$.

	\textbf{Remark.} In the proposed meta-learning objective, the meta-test loss encourages the loss of meta-train samples to optimize the model towards a direction that can improve the accuracy of meta-test samples. By iteratively enforcing the generalization process from meta-train domains to meta-test domain, the model can avoid overfitting to domain bias and can learn domain-invariant representations that generalize well on unseen domains.
	
	Next, we will introduce the loss functions used in meta-learning in Sec.~\ref{sec:memory}, the MetaBN layer in Sec.~\ref{sec:metabn} and detailed training procedure of meta-learning in Sec.~\ref{sec:procedures}.

	\subsection{Memory-based Identification Loss}
	\label{sec:memory}
	
	Identification loss can effectively learn discriminative person representations in a classification manner. Commonly, a fully-connected layer is adopted as the classifier to produce the probabilities that are used for computing the cross-entropy loss. Although existing works~\cite{balaji2018metareg,han2018coteaching,sun2019meta} show the effectiveness of meta-learning in the classification task, the parametric classifier is inadequate in the context of ReID. This is because ReID is an open-set task, where different domains contain completely different identities and the number of identities in each domain is commonly large. In multi-source DG of ReID, we have two kinds of parametric classifier selections, one global FC classifier or $N_S$ parallel FC classifiers for each domain, both of which will lead to problems during meta-learning.

\begin{figure}[!t]
		\centering
		\includegraphics[width=0.5\textwidth]{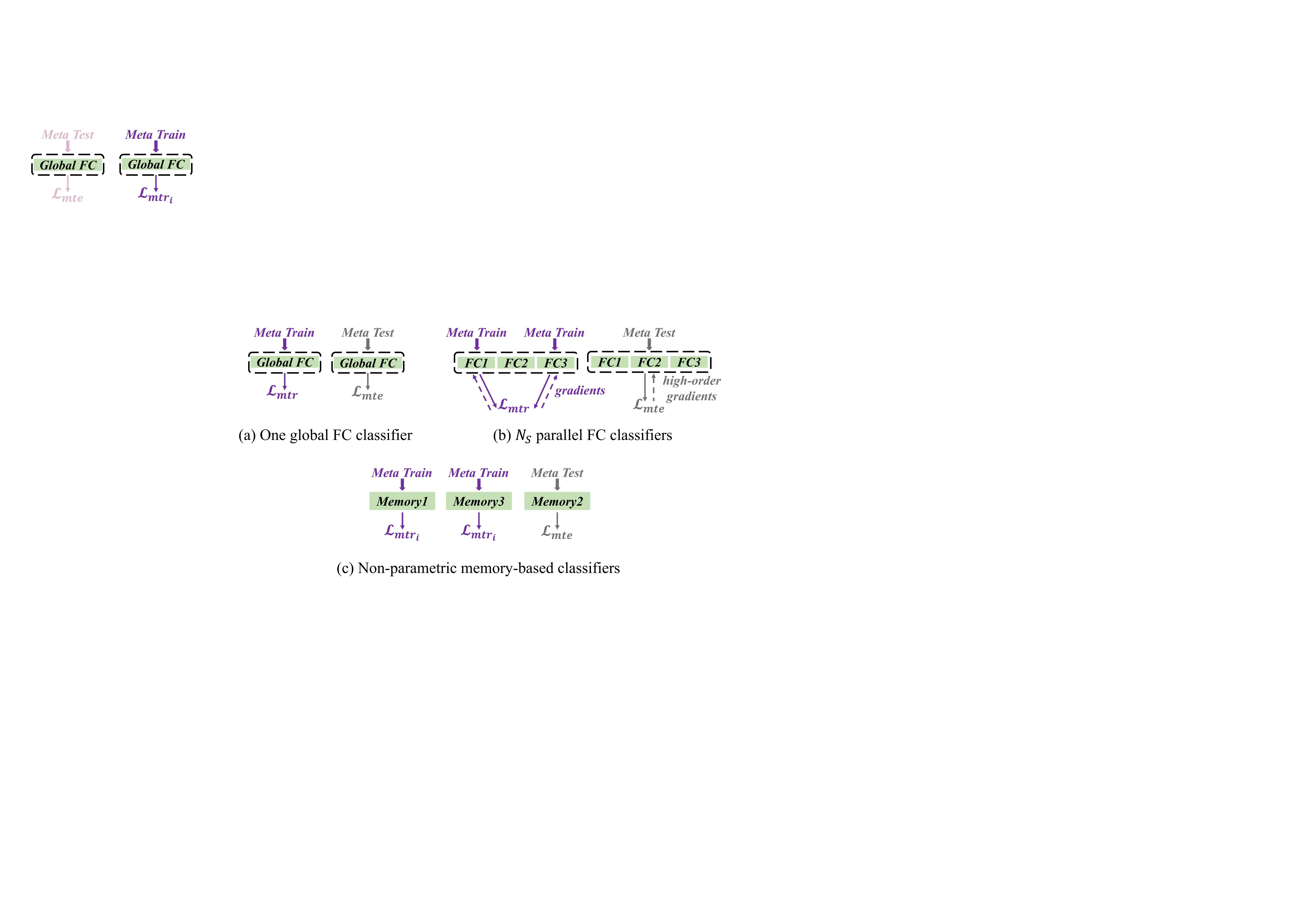}
		\vspace{-.2in}
		\caption{Comparison of different classifiers. Layers within dashed rectangles are updated together by an optimizer.}
		\label{fig:classifier}
	\end{figure}

    For the global FC classifier (Fig.~\ref{fig:classifier}(a)), the dimension of the FC layer is the sum of all source identities. Different from closed-set tasks~\cite{balaji2018metareg,2020EccvDMG}, the global FC classifier contains a large number of parameters when trained with multiple person ReID datasets. This will lead to unstable optimization during the meta-learning. As for parallel FC classifiers in Fig.~\ref{fig:classifier}(b), although we can alleviate the parameter burden by only identifying persons within their own domain classifier, the number of parameters for all classifiers is still large. Moreover, during the meta-learning, the classifier of the meta-test domain is only updated by high-order gradients, which is asynchronous with the feature encoder. This optimization process is unequal and unstable, leading to an incomplete usage of meta-learning.

	Taking all the above into consideration, inspired by~\cite{ge2020self,memory,zhong2019invariance,zhong2020memory}, we propose a memory-based identification loss for multi-source DG, which is non-parametric and suitable for both meta-learning and person ReID. 
	As shown in Fig.~\ref{fig:classifier}(c), we maintain a feature memory for each domain, which contains the centroids of each identity. The similarities between features and memory centroids are used to compute the identification loss. The memory-based identification loss has two advantages to our meta-learning framework. \textit{First}, the memory is a non-parametric classifier, which avoids the unstable optimization caused by a large number of parameters. \textit{Second}, the asynchronous update between the feature encoder and memory has a slight influence on model training. This is because the memory is updated smoothly by a momentum instead of being updated by an optimizer. Thus, the memory is insensitive to the changes of the feature encoder caused by the last few training iterations. In Sec.~\ref{sec:ablation}, we show that our meta-learning framework gains more improvements with the memory-based identification loss than with the FC-based identification loss. Next, we will introduce the memory-based identification loss in detail. 
	\par
	\textbf{Memory Initialization.} We maintain an individual memory for each source domain. For a source domain $\mathcal{D}^i_S$ with $n_i$ identities, the memory $\mathcal{M}^i$ has $n_i$ slots, where each slot saves the feature centroid of the corresponding identity. In initialization, we use the model to extract features for all samples of $\mathcal{D}_S^i$. Then, we initialize the centroid of each identity with a feature, which is averaged on the features of the corresponding identity. For simplicity, we omit the superscript of the domain index and introduce the memory updating and memory-identification loss for one domain.
	\par
	\textbf{Memory Updating.} At each training iteration, we update the memory with the features in the current mini-batch. A centroid in the memory is updated through,
	\begin{equation}
	\mathcal{M}[k] \gets m\cdot \mathcal{M}[k] + (1-m)\cdot \frac{1}{|\mathcal{B}_k|}\sum\limits_{x_i\in \mathcal{B}_k}f(x_i) ,
	\end{equation}
	where $\mathcal{B}_k$ denotes the samples belonging to the $k$th identity and $|\mathcal{B}_k|$ denotes the number of samples for the $k$th identity in current mini-batch. $m\in [0,1]$ controls the updating rate.
	\par
	\textbf{Memory-based identification loss.} Given an embedding feature $f(x_i)$ from the forward propagation, we calculate the similarities between $f(x_i)$ and each centroid in the memory. The memory-based identification loss aims to classify $f(x_i)$ into its own identity, which is calculated by:
	\begin{equation}
	\label{loss:identification}
	\mathcal{L}_M = -\log{\frac{\exp{\left(\mathcal{M}[i]^T f(x_i) / \tau \right)}}{\sum_{k=1}^{n_i}\exp{\left(\mathcal{M}[k]^T f(x_i) / \tau \right)}}} ,
	\end{equation} 
	where $\tau$ is the temperature factor that controls the scale of distribution.
		
	\textbf{Triplet loss.} We also use triplet loss~\cite{triplet} to train the model, which is formulated as,
	\begin{equation}
	\label{loss:triplet}
	\mathcal{L}_{Tri} = [d_p - d_n + \delta]_+ ,
	\end{equation}
	where $d_p$ is the Euclidean distance between an anchor feature and a hard positive feature, and $d_n$ is the Euclidean distance between an anchor feature and a hard negative feature. $\delta$ is the margin of triplet loss and $[\cdot]_+$ refers to $max(\cdot,0)$. 
	
	\subsection{MetaBN}
	\label{sec:metabn}
		
	\begin{figure}[t]
	    \centering
	    \includegraphics[width=0.45\textwidth]{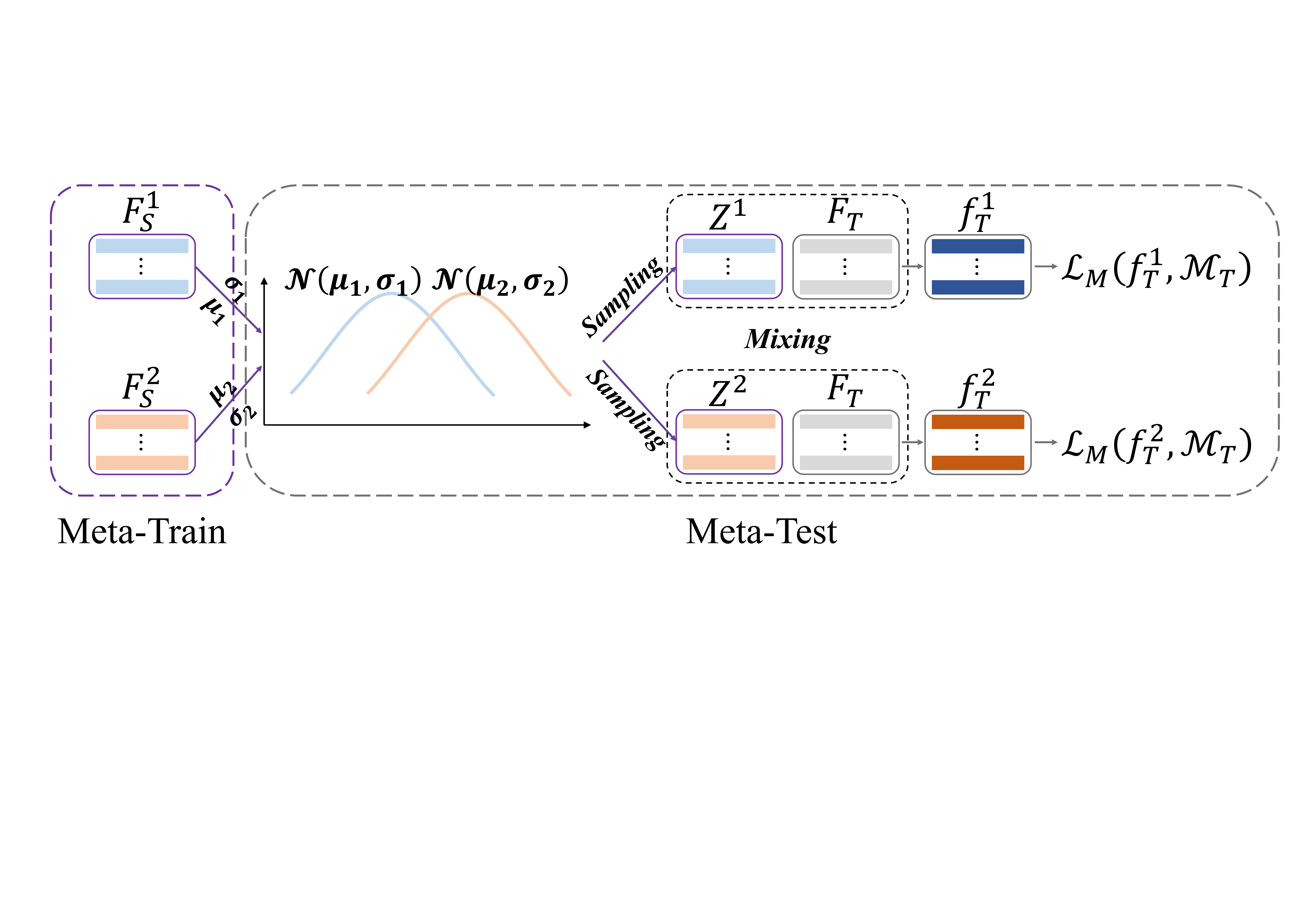}
	    \vspace{-.1in}
	    \caption{Detailed architecture of MetaBN. MetaBN first utilizes the mean and variance of meta-train mini-batch features to construct Gaussian Distributions. Then, features sampled from these distributions are mixed with meta-test mini-batch features for generating new meta-test features.}
	    \label{fig:MetaBN}
	\end{figure}
	In our meta-learning strategy, the meta-test loss is important for learning generalizable representations, since the meta-test plays the role of the ``unseen'' domain. Intuitively, if the meta-test examples are sampled from more diverse distributions, the model will be optimized to be more robust to variations and thus be more generalizable to unseen domains. To achieve this goal, we introduce MetaBN to generate more diverse meta-test features at the feature-level. As shown in Fig.~\ref{fig:overall}, we replace the last batch normalization layer (BN)~\cite{bn} in the network with MetaBN. During training, MetaBN utilizes the domain information from meta-train domains to inject domain-specific information into meta-test features. This process can diversify meta-test features, enabling the model to simulate more feature variations. The operation of MetaBN is illustrated in Fig.~\ref{fig:MetaBN}.
	\par
	In the meta-train stage, for the $i$th meta-train domain, MetaBN normalizes the meta-train features as the traditional BN, and saves the mini-batch mean $\mu_i$ and mini-batch variance $\sigma_i$, which are used in the following meta-test stage. 
	
	In the meta-test stage, MetaBN uses the saved mean and variance to form  $N_S-1$ Gaussian Distributions. Note that, the generated distribution mainly reflects the high-level domain information instead of specific identity information. This is because each saved mean and variance is calculated over samples belonging to dozens of identities. Considering this factor, we sample features from these distributions and inject these domain-specific features into meta-test features.

	Specifically, for the $i$th distribution, we sample one feature $z^i_j$ for each meta-test feature:
	\begin{equation}
	    z^i_j \sim \mathcal{N}\left(\mu_i, \sigma_i \right),
	\end{equation}
	where $\mathcal{N}$ denotes Gaussian Distribution. By doing so, we obtain $B$ (the batch size of meta-test features) sampled features, which are mixed with the original meta-test features for generating new features $F_{T}^{i}$,
	\begin{equation}
	   F_{T}^{i} = \lambda F_T + (1-\lambda) Z^i,
	\end{equation}
	where $F_T$ denotes the original meta-test features. $Z^i = [z^i_0, z^i_1, \cdots, z^i_B]$ denotes $B$ sampled features from the $i$th Gaussian Distribution.
	$\lambda$ is the mixing coefficient, which is sampled from Beta Distribution, \ie, $\lambda \sim \rm{Beta}(1,1)$. 

	Finally, the mixed features are normalized by batch normalization,
	\begin{equation}
	   f_{T}^i = \gamma \frac{F_{T}^{i}-\mu_{T}^{i}}{\sqrt{{\sigma_{T}^{i}}^2 + \epsilon}} + \beta,
	\end{equation}
	where $\mu_{T}^{i}$ and ${\sigma_{T}^{i}}$ denote mini-batch mean and variance of $F_{T}^{i}$. $\gamma$ and $\beta$ denote the learnable parameters that scale and shift the normalized value.

	\subsection{Training procedure of M$^3$L}
	\label{sec:procedures}
	
	During training, $N_S$ source domains are separated into $N_S-1$ meta-train domains and \emph{one} meta-test domain at each iteration. The model is optimized by the losses calculated in the meta-train and meta-test stages.
	
	\begin{algorithm}[t]
		\caption{Training procedure of M$^3$L}
		\label{alg:metalearning}
		\SetKwInput{Kwinit}{Init}{}{}
		\SetKwInput{kwInput}{Input}{}{}
		\kwInput{$N_S$ source domains $\mathcal D_S = \{\mathcal D^1_S, ...,\mathcal D^{N_S}_S\}$}	
		\Kwinit{Feature encoder $F$ parametrized by $\Theta$; \qquad
			Inner loop learning rate $\alpha$; \qquad \qquad \qquad \qquad
			Outer loop learning rate $\beta$; 
			Batch-size $B$;} 
			
			\For{iter in train\_iters}{
				Randomly select a domain as meta-test $D_{mte}$; \\ 
				Sample remaining domains as meta-train $D_{mtr}$; \\
				\textbf{Meta-Train:}\\
				Sample $B$ images $X_S^k$ from each meta-train domain $D^k_{mtr}$; \\
				Compute meta-train loss $\mathcal{L}_{mtr}$ (Eq.\ref{loss:totalmtr});\\
				Copy the original model and update the copied parameters $\Theta$ by Adam and inner loop learning rate $\alpha$:\\
				$\Theta^{'} \leftarrow Adam(\nabla_{\Theta}\mathcal{L}_{mtr}, \Theta, \alpha)$;\\
				\textbf{Meta-Test:}\\
				Sample $B$ images $X_T$ from the meta-test domain $D_{mte}$;\\
				Diversify meta-test features with MetaBN; \\
				Compute meta-test loss $\mathcal{L}_{mte}$ (Eq.\ref{loss:mte}); \\ 
				\textbf{Meta Optimization:}\\
				Update the original model parameters $\Theta$ by Adam and outer loop learning rate $\beta$: \\
				Gradient: $g \leftarrow \nabla_{\Theta}(\mathcal{L}_{mtr}(\Theta) + \mathcal{L}_{mte}(\Theta^{'}))$;\\
				$\Theta \leftarrow Adam(g,\Theta,\beta)$.\\
			}
	\end{algorithm}
	
	\textbf{Meta-train.} For each meta-train domain, the meta-train loss is a combination of memory-based identification (Eq.\ref{loss:identification}) and triplet losses (Eq.\ref{loss:triplet}), \ie,
	\begin{equation}
	\label{loss:mtr}
	\mathcal{L}_{mtr}^i = \mathcal{L}_{Tri}(X^i_S;\Theta) + \mathcal{L}_{M}(X^i_S, \mathcal{M}^i_S;\Theta),
	\end{equation}
	where $\Theta$ denotes the parameters of the network. $X^i_S$ and $\mathcal{M}^i_S$ denote the training samples and memory of the $i$th meta-train domain, respectively.
	
	The total loss for meta-train is averaged over $N_S-1$ meta-train domains, formulated as,
	\begin{equation}
	\label{loss:totalmtr}
	\mathcal{L}_{mtr} = \frac{1}{N_S-1}\sum\limits_{i=0}^{N_S-1}\mathcal{L}_{mtr}^i.
	\end{equation}
	
	\textbf{Meta-test.} In the meta-test stage, the meta-test domain is performed on the new parameters $\Theta^{'}$, which is obtained by optimizing $\Theta$ with $\mathcal{L}_{mtr}$. With the MetaBN proposed in Sec.~\ref{sec:metabn}, we can obtain $N_S-1$ mixed features for each meta-test sample. The average memory-based identification loss over these features is considered as the meta-test memory-based identification loss. The meta-test loss is:
	\begin{equation}
	\small
	\label{loss:mte}
	\begin{aligned}
	\mathcal{L}_{mte} &= \mathcal{L}_{Tri}(X_T;\Theta^{'})+\frac{1}{N_S-1}\sum\limits_{k=0}^{N_S-1}\mathcal{L}_M(f_{T}^{k}, \mathcal{M}_T;\Theta^{'}),   
	\end{aligned}
	\end{equation}
	where $X_T$ denotes the meta-test samples and $f_{T}^{k}$ denotes the $k$th mixed features generated by the MetaBN.
	
	\textbf{Meta Optimization.} Finally, the model is optimized by the objective in Eq.\ref{loss:meta}. The optimization procedure of our M$^3$L is summarized in Alg.~\ref{alg:metalearning}.

	\section{Experiments}
	\label{sec:experiments}
	\subsection{Benchmarks and Evaluation Metrics}
	We conduct experiments on four large-scale person re-identification benchmarks: Market-1501~\cite{market1501}, DukeMTMC-reID~\cite{dukemtmc2,dukemtmc}, CUHK03~\cite{cuhk03,rerank} and MSMT17~\cite{msmt17}. \textit{For studying the multi-source DG}, we divide these four datasets into two parts: three domains as source domains for training and the other one as target domain for testing. 
	The statistics of these four benchmarks are shown in Table \ref{tab:datasets}. 
	For simplicity, we denote Market-1501, DukeMTMC-reID, CUHK03, and MSMT17 as M, D, C, and MS in tables.
	\begin{table}[t]
		\caption{Statistics of Person ReID Benchmarks.}
		\centering
		\label{tab:datasets}
		\vspace{-3mm}
		\fontsize{9pt}{10pt}\selectfont
		\begin{tabular}{p{3cm}|p{1cm}<{\centering}p{1.2cm}<{\centering}p{1.3cm}<{\centering}}
			\toprule
			Benchmarks & \# IDs & \# images & \# cameras \\
			\midrule
			Market-1501~\cite{market1501} & 1,501 & 32,217 & 6  \\
			DukeMTMC-reID~\cite{dukemtmc} & 1,812 & 36,411 & 8 \\
			CUHK03~\cite{cuhk03} & 1,467 & 28,192 & 2 \\
			MSMT17~\cite{msmt17} & 4,101 & 126,441 & 15 \\
			\bottomrule		
		\end{tabular}
	\end{table}

	\textit{\textbf{Note:} In default, for CUHK03, we use the old protocol (CUHK03, 26,263 images of 1,367 IDs for training) as the source domain for training the model and the detected subset of the new protocol (CUHK-NP~\cite{rerank}) as the target domain for testing; for MSMT17, we use the MSMT17\_V2 for both training and testing.
	In Table~\ref{tab:sota}, we also provide the results of using the detected subset of CUHK-NP (7,365 images of 767 IDs for training) and MSMT17\_V1 for both training and testing, and we recommend using this setting in future studies.}
	
	The cumulative matching characteristic (CMC) at Rank-1 and mean average precision (mAP) are used to evaluate performance on the target testing set.
	
	\subsection{Implementation Details}
	We implement our method with two common backbones, \textit{i.e.}, ResNet-50~\cite{resnet} and IBN-Net50~\cite{ibn}. 
	Images are resized to 256$\times$128 and the training batch size is set to 64. We use random flipping and random cropping for data augmentation. For the memory, the momentum coefficient $m$ is set to 0.2 and the temperature factor $\tau$ is set to 0.05. The margin $\delta$ of triplet loss is 0.3. To optimize the model, we use Adam optimizer with a weight decay of 0.0005. The learning rate of inner loop $\alpha$ and outer loop $\beta$ are initialized to $3.5\times10^{-5}$ and increase linearly to $3.5\times10^{-4}$ in the first 10 epochs. Then, $\alpha$ and $\beta$ are decayed by 0.1 at the 30th epoch and 50th epoch. The total training stage takes 60 epochs. 
	
	\textbf{Baseline.} For the baseline, we directly train the model with the memory-based identification loss and triplet loss using the data of all the source domains. That is, the baseline does not apply the meta-learning strategy and MetaBN. 
	
	\subsection{Comparison with State-of-the-Art methods}

\begin{table*}[t]
		\caption{Comparison with State-of-the-Arts domain generalization methods on four large-scale person ReID benchmarks --- Market-1501 (M), DukeMTMC-reID (D), CUHK03 (C) and MSMT17 (MS). 
		The performance is evaluated quantitatively by mean average precision (mAP) and cumulative matching characteristic (CMC) at Rank-1 (R1).}
		\vspace{-.15in}
		\centering
		\label{tab:sota}
		\fontsize{8.5pt}{9.5pt}\selectfont
		\begin{threeparttable}
			\begin{tabular}{p{2.4cm}|p{1.3cm}<{\centering}|p{1cm}<{\centering}|p{1cm}<{\centering}|p{0.8cm}<{\centering}p{0.8cm}<{\centering}|p{1.3cm}<{\centering}|p{1cm}<{\centering}|p{1cm}<{\centering}|p{0.8cm}<{\centering}p{0.8cm}<{\centering}}
				\toprule
				\multirow{2}{*}{Method} & \multirow{2}{*}{Source} & \multirow{2}{*}{IDs} & \multirow{2}{*}{Images} & \multicolumn{2}{c|}{Market-1501}& \multirow{2}{*}{Source} & \multirow{2}{*}{IDs} & \multirow{2}{*}{Images} & \multicolumn{2}{c}{DukeMTMC} \\
				&&&& mAP & R1 &&&& mAP & R1\\
				\midrule
				OSNet-IBN~\cite{zhou2019omni} & \multirow{4}{*}{Com-MS} & \multirow{4}{*}{4,101} & \multirow{4}{*}{126,441} & 37.2 & 66.5 & \multirow{4}{*}{Com-MS} & \multirow{4}{*}{4,101} & \multirow{4}{*}{126,441} & 45.6 & 67.4 \\
				OSNet-AIN~\cite{zhou2019learning} &  &  &  & 43.3 & 70.1 & &  &  & 52.7 & 71.1 \\
				SNR~\cite{jin2020style} &  &  &  & 41.4 & 70.1 &  &  &  & 50.0 & 69.2 \\
				QAConv$_{50}$~\cite{Liao2020QAConv} &  &  &  & 43.1 & 72.6 &  &  &  & 52.6 & 69.4 \\
				\midrule
				QAConv$_{50}$~\cite{Liao2020QAConv}* & \multirow{3}{*}{MS+D+C} & \multirow{3}{*}{3,110} & \multirow{3}{*}{75,406} & 35.6 & 65.7 & \multirow{3}{*}{MS+M+C} & \multirow{3}{*}{3,159} & \multirow{3}{*}{71,820} & 47.1 & 66.1 \\
				M$^3$L~(ResNet-50) &  &  &  & 48.1 & 74.5 &  &  &  & 50.5 & \textbf{69.4} \\
				M$^3$L~(IBN-Net50) &  &  &  & \textbf{50.2} & \textbf{75.9} &  &  &  & \textbf{51.1} & 69.2 \\
				\midrule
				QAConv$_{50}$~\cite{Liao2020QAConv}*\dag & \multirow{3}{*}{\shortstack{MS+D\\+C-NP}} &\multirow{3}{*}{2,510} & \multirow{3}{*}{56,508} & 39.5 & 68.6 & \multirow{3}{*}{\shortstack{MS+M\\+C-NP}}& \multirow{3}{*}{2,559} & \multirow{3}{*}{52,922} & 43.4 & 64.9 \\
				M$^3$L~(ResNet-50)\dag &  &  &   & 51.1 & 76.5 &  &  &  & 48.2 & 67.1 \\
				M$^3$L~(IBN-Net50)\dag &  &  &  & \textbf{52.5} & \textbf{78.3} &  &  &  & \textbf{48.8} & \textbf{67.2} \\
				\bottomrule
				\toprule
				\multirow{2}{*}{Method} & \multirow{2}{*}{Source} & \multirow{2}{*}{IDs} & \multirow{2}{*}{Images} & \multicolumn{2}{c|}{CUHK-NP}& \multirow{2}{*}{Source} & \multirow{2}{*}{IDs} & \multirow{2}{*}{Images} & \multicolumn{2}{c}{MSMT17} \\ 
				&&&& mAP & R1 &&&& mAP & R1\\
				\midrule
				QAConv$_{50}$~\cite{Liao2020QAConv} & Com-MS & 4,101 & 126,441 & 22.6 & 25.3 & D & 702 & 16,522 & 8.9 & 29.0 \\
				\midrule
				QAConv$_{50}$~\cite{Liao2020QAConv}* & \multirow{3}{*}{MS+D+M} & \multirow{3}{*}{2,494} & \multirow{3}{*}{62,079} & 21.0 & 23.5 & \multirow{3}{*}{D+M+C} & \multirow{3}{*}{2,820} & \multirow{3}{*}{55,748} & 7.5 & 24.3 \\
				M$^3$L~(ResNet-50) &  &  &  & 29.9 & 30.7 &  &  &  & 12.9 & 33.0 \\
				M$^3$L~(IBN-Net50) &  &  &  & \textbf{32.1} & \textbf{33.1} &  &  &  & \textbf{14.7} & \textbf{36.9} \\
				\midrule
				QAConv$_{50}$~\cite{Liao2020QAConv}*\dag & \multirow{3}{*}{MS+D+M} &\multirow{3}{*}{2,494} & \multirow{3}{*}{62,079} & 19.2 & 22.9 & \multirow{3}{*}{\shortstack{D+M\\+C-NP}}& \multirow{3}{*}{2,220} & \multirow{3}{*}{36,823} & 10.0 & 29.9 \\
				M$^3$L~(ResNet-50)\dag &  & & & 30.9 & \textbf{31.9} &  &  &  & 13.1 & 32.0 \\
				M$^3$L~(IBN-Net50)\dag &  &  &  & \textbf{31.4} & 31.6 &  &  &  & \textbf{15.4} & \textbf{37.1} \\
				\bottomrule
				
			\end{tabular}
			\begin{tablenotes}
				\scriptsize
				\item[*] We reimplement this work based on the authors' code on Github with the same source datasets as us. \\
				\item[\dag] Model is trained and tested on the detected subset of CUHK-NP and MSMT17\_V1.
			\end{tablenotes}
		\end{threeparttable}
	\end{table*}
	
	Since there is no multi-source DG method evaluating on large-scale datasets, we compare our method with state-of-the-art single-source DG methods,
	including OSNet-IBN~\cite{zhou2019omni}, OSNet-AIN~\cite{zhou2019learning}, SNR~\cite{jin2020style} and QAConv~\cite{Liao2020QAConv}. SNR~\cite{jin2020style} and QAConv~\cite{Liao2020QAConv} use the ResNet-50 as the backbone. OSNet-IBN~\cite{zhou2019omni} and OSNet-AIN~\cite{zhou2019learning} use their self-designed networks that have better performance than ResNet-50. When testing on Market-1501, DukeMTMC-reID, and CUHK03, the existing single-source DG methods utilize MSMT17 as the source domain for model training. They combine the train set and test set of MSMT17, which is denoted as Combined MSMT17 (Com-MS) in this paper. 
	To verify that the effectiveness of our method is obtained by multi-source meta-learning instead of training with more IDs and images, we only use the training sets of the source domains for model training. 
	For example, when using Market-1501 as the target domain, we train the model with the train sets of DukeMTMC-reID, CUHK03, and MSMT17, including 3,110 IDs and 75,406 images. The numbers of IDs and images are less than that of Combined MSMT17 (3,110 IDs \vs 4,101 IDs, and, 75,406 images \vs 126,441 images). To conduct a fair comparison, we reimplement recent published QAConv~\cite{Liao2020QAConv} with the same training data as us.
	Comparison results are reported in Table \ref{tab:sota}.

	\textbf{Results on Market-1501 and DukeMTMC-reID.} From Table \ref{tab:sota}, we can make the following observations. 
	First, when using Combined MSMT17 as the source data, OSNet-AIN~\cite{zhou2019learning} and QAConv~\cite{Liao2020QAConv} achieve the best results on both Market-1501 and DukeMTMC-reID. 
	Second, compared to single-source DG methods that use more training data (Combined MSMT17), our M$^3$L outperforms them by a large margin on Market-1501 and achieves comparable results with them on DukeMTMC-reID. Specifically, when testing on Market-1501, with the same backbone, our M$^3$L surpasses SNR~\cite{jin2020style} by 6.7\% in mAP and 4.4\% in Rank-1 accuracy. 
	Third, when training with multiple source domains, with the same backbone, our M$^3$L produces significantly higher results than QAConv$_{50}$. Specifically, our M$^3$L is higher than QAConv$_{50}$ by 12.5\% in mAP for Market-1501 and by 3.4\% in mAP for DukeMTMC-reID. This demonstrates the superiority of our method over the method that considers all the source domains as one domain.
	Fourth, when using the IBN-Net50 as the backbone, our M$^3$L can achieve better mAP than using ResNet-50.

	\textbf{Results on CUHK03 and MSMT17.} There is only one method (QAConv~\cite{Liao2020QAConv}) evaluated on CUHK03 and MSMT17. When testing on MSMT17,  QAConv~\cite{Liao2020QAConv} uses DukeMTMC-reID as the source data. Clearly, our M$^3$L achieves higher results than QAConv~\cite{Liao2020QAConv} on both datasets, no matter how many source domains QAConv is trained with.
	We also find that both our M$^3$L and QAConv produce poor results on CUHK03 and MSMT17, indicating there is still a large room for generalizable models in DG.

	\subsection{Ablation Studies}
	\label{sec:ablation}
	
	\textbf{Effectiveness of Meta-Learning.} To investigate the effectiveness of the proposed meta-learning strategy, we conduct ablation studies in Table \ref{tab:metalearning}. Clearly, the model trained with the proposed meta-learning strategy consistently improves the results with different backbones.
	Specifically, with ResNet-50, adding meta-learning optimization increases the baseline by 5.3\% in Rank-1 accuracy on Market-1501 and by 3.7\% in Rank-1 accuracy on CUHK03. 
	With IBN-Net50 backbone, meta-learning strategy gains 5.4\% and 2.8\% improvement in mAP on Market-1501 and CUHK03, respectively.
	This demonstrates that by simulating the train-test process during training, the meta-learning strategy helps the model to learn domain-invariant representations that can perform well on unseen domains.

	\textbf{Effectiveness of MetaBN.}
	As shown in Table~\ref{tab:metalearning}, plugging MetaBN into the meta-learning-based model further improves the generalization ability. 
	For ResNet-50 backbone, MetaBN improves the meta-optimized model by 1.3\% and 1.6\% in Rank-1 accuracy on Market-1501 and CUHK03. For IBN-Net50, we can observe similar improvements.
	The results validate that diversifying meta-test features by MetaBN is able to help the model to learn more generalizable representations for unseen domains.

    \begin{table}[t]
	    \centering
	    \caption{Ablation studies on meta-learning strategy and MetaBN. Models are trained with the other three datasets except the target dataset. ``Meta'': training with meta-learning strategy; ``MetaBN'': training with MetaBN.}
	    \label{tab:metalearning}
	    \vspace{-3mm}
	    \fontsize{8pt}{8pt}\selectfont
	    \begin{tabular}{p{1.3cm}|p{0.5cm}<{\centering}|p{0.9cm}<{\centering}|p{0.6cm}<{\centering}p{0.6cm}<{\centering}|p{0.6cm}<{\centering}p{0.6cm}<{\centering}}
	         \toprule
				\multirow{2}{*}{Backbone} & \multirow{2}{*}{Meta} & \multirow{2}{*}{MetaBN} & \multicolumn{2}{c|}{MS+D+C$\rightarrow$M}
				& \multicolumn{2}{c}{MS+D+M$\rightarrow$C} \\ 
				&&& mAP & R1 & mAP & R1\\
				\midrule
	            \multirow{3}{*}{ResNet-50} & $\times$& $\times$ &  41.1 & 67.9 & 25.7 & 25.4 \\
	            & \checkmark & $\times$ & 47.4 & 73.2 & 29.1 & 29.1 \\
	            & \checkmark & \checkmark & \textbf{48.1} & \textbf{74.5} & \textbf{29.9} & \textbf{30.7} \\
	            \midrule
	            \multirow{3}{*}{IBN-Net50} & $\times$& $\times$ & 43.6 & 71.1 & 28.2 & 29.4 \\
	             & \checkmark& $\times$ & 49.0 & 75.0 & 31.0 & 31.8 \\
	             & \checkmark& \checkmark & \textbf{50.2} & \textbf{75.9} & \textbf{32.1} & \textbf{33.1} \\
	            \bottomrule
	    \end{tabular}
	\end{table}
	
\textbf{Loss function components.} We conduct experiments to evaluate the impact of the memory-based identification loss and triplet loss. Results in Table \ref{tab:loss} show that the memory-based identification loss $\mathcal{L}_M$ is the predominant supervision for training a generalizable model and additionally adding the triplet loss $\mathcal{L}_{Tri}$ can slightly improve the performance.

\textbf{Comparison of different classifiers.}  In Table~\ref{tab:idloss}, we compare different types of identification classifiers.
We have the following observations.
First, compared with the two parametric classifiers, our proposed non-parametric classifier gains higher improvement with the meta-learning strategy. Second, when directly training with multi-source data without the meta-learning, the model trained with memory-based identification loss achieves higher results. These two observations demonstrate that the proposed memory-based identification loss is suitable for multi-source DG and our meta-learning strategy. 

\begin{table}[t]
\centering
\caption{Comparison of loss function components. $\mathcal{L}_M$ and $\mathcal{L}_{Tri}$ denote memory-based identification loss and triplet loss. Experiments are conducted with ResNet-50.}
\label{tab:loss}
\vspace{-3mm}
\fontsize{8pt}{9pt}\selectfont
\begin{tabular}{p{0.7cm}<{\centering}|p{0.7cm}<{\centering}|p{0.8cm}<{\centering}p{0.8cm}<{\centering}|p{0.8cm}<{\centering}p{0.8cm}<{\centering}}
     \toprule
		\multirow{2}{*}{$\mathcal{L}_M$} & \multirow{2}{*}{$\mathcal{L}_{Tri}$} & \multicolumn{2}{c|}{MS+D+C$\rightarrow$M}
		& \multicolumn{2}{c}{MS+D+M$\rightarrow$C} \\ 
		&& mAP & R1 & mAP & R1\\
		\midrule
        \checkmark& $\times$ & 47.8 & 73.6 & \textbf{29.9} & 30.3 \\
        $\times$ & \checkmark & 35.1 & 59.7 & 20.9 & 19.9 \\
       \checkmark & \checkmark & \textbf{48.1} & \textbf{74.5} & \textbf{29.9} & \textbf{30.7} \\
        \bottomrule
\end{tabular}
\end{table}
	
\begin{table}[t]
\centering
\caption{Comparison of different classifiers. $\mathcal{L}_M$ denotes memory-based identification loss. $\mathcal{L}_{FCG}$ and $\mathcal{L}_{FCP}$ denote traditional identification loss with global classifier and parallel classifiers. ``Meta'' denotes training with the meta-learning strategy and MetaBN. Numbers in parentheses denote performance improvement gained by ``Meta''. Experiments are conducted with ResNet-50.}
\label{tab:idloss}
\vspace{-3mm}
\fontsize{8pt}{9pt}\selectfont
\begin{tabular}{p{1cm}|p{0.5cm}<{\centering}|p{1cm}<{\centering}p{1cm}<{\centering}|p{1cm}<{\centering}p{1cm}<{\centering}}
     \toprule
		\multirow{2}{*}{Loss} & \multirow{2}{*}{Meta} & \multicolumn{2}{c|}{MS+D+C$\rightarrow$M}
		& \multicolumn{2}{c}{MS+D+M$\rightarrow$C} \\
		& & mAP & R1 & mAP & R1\\
		\midrule
        \multirow{2}{*}{$\mathcal{L}_{FCG}$} & $\times$ & 37.7 & 67.0 &  21.2 & 20.9 \\
        & \checkmark & 39.7~(\textcolor{red}{2.0}) & 68.3~(\textcolor{red}{1.3}) & 21.2~(\textcolor{red}{0.0}) & 21.9~(\textcolor{red}{1.0}) \\
        \midrule
        \multirow{2}{*}{$\mathcal{L}_{FCP}$} & $\times$ & 37.7 & 67.0 & 21.2 & 20.9 \\
        & \checkmark & 40.9~(\textcolor{red}{3.2}) & 69.3~(\textcolor{red}{2.3}) & 23.9~(\textcolor{red}{2.7})  & 24.3~(\textcolor{red}{3.4}) \\
        \midrule
        \multirow{2}{*}{$\mathcal{L}_{M}$} & $\times$ & 41.1 & 67.9 & 25.7 & 25.4 \\
        & \checkmark & 48.1~(\textbf{\textcolor{red}{7.0}}) & 74.5~(\textbf{\textcolor{red}{6.6}}) & 29.9~(\textbf{\textcolor{red}{4.2}}) & 30.7~(\textbf{\textcolor{red}{5.3}}) \\
        \bottomrule
\end{tabular}
\end{table}

\textbf{Effectiveness of Multi-Source.}
Table \ref{tab:source} shows the comparison between two-source DG and multi-source DG. Despite bringing more domain bias, training with more source domains consistently produces higher results when testing on an unseen domain. This demonstrates the significance of studying multi-source DG.
	
\begin{table}[t]
\centering
\caption{Comparison of training with different source domains. Experiments are conducted with ResNet-50.}
\label{tab:source}
\vspace{-3mm}
\fontsize{8pt}{9pt}\selectfont
\begin{tabular}{p{1.2cm}|p{0.7cm}<{\centering}p{0.7cm}<{\centering}|p{1.2cm}|p{0.7cm}<{\centering}p{0.7cm}<{\centering}}
     \toprule
		\multirow{2}{*}{Sources} & \multicolumn{2}{c|}{Market-1501}
		& \multirow{2}{*}{Sources} & \multicolumn{2}{c}{CUHK03} \\ 
		& mAP & R1 & & mAP & R1\\
		\midrule
        MS+D & 38.5  & 66.2 & MS+D & 21.9 & 23.7 \\
        MS+C & 39.8 & 65.8 & MS+M & 27.1 & 27.8 \\
        MS+D+C & \textbf{48.1} & \textbf{74.5} & MS+D+M & \textbf{29.9} & \textbf{30.7} \\
        \bottomrule
\end{tabular}
\end{table}

\subsection{Visualization}
	\begin{figure}[t]
		\centering
		\includegraphics[width=0.5\textwidth]{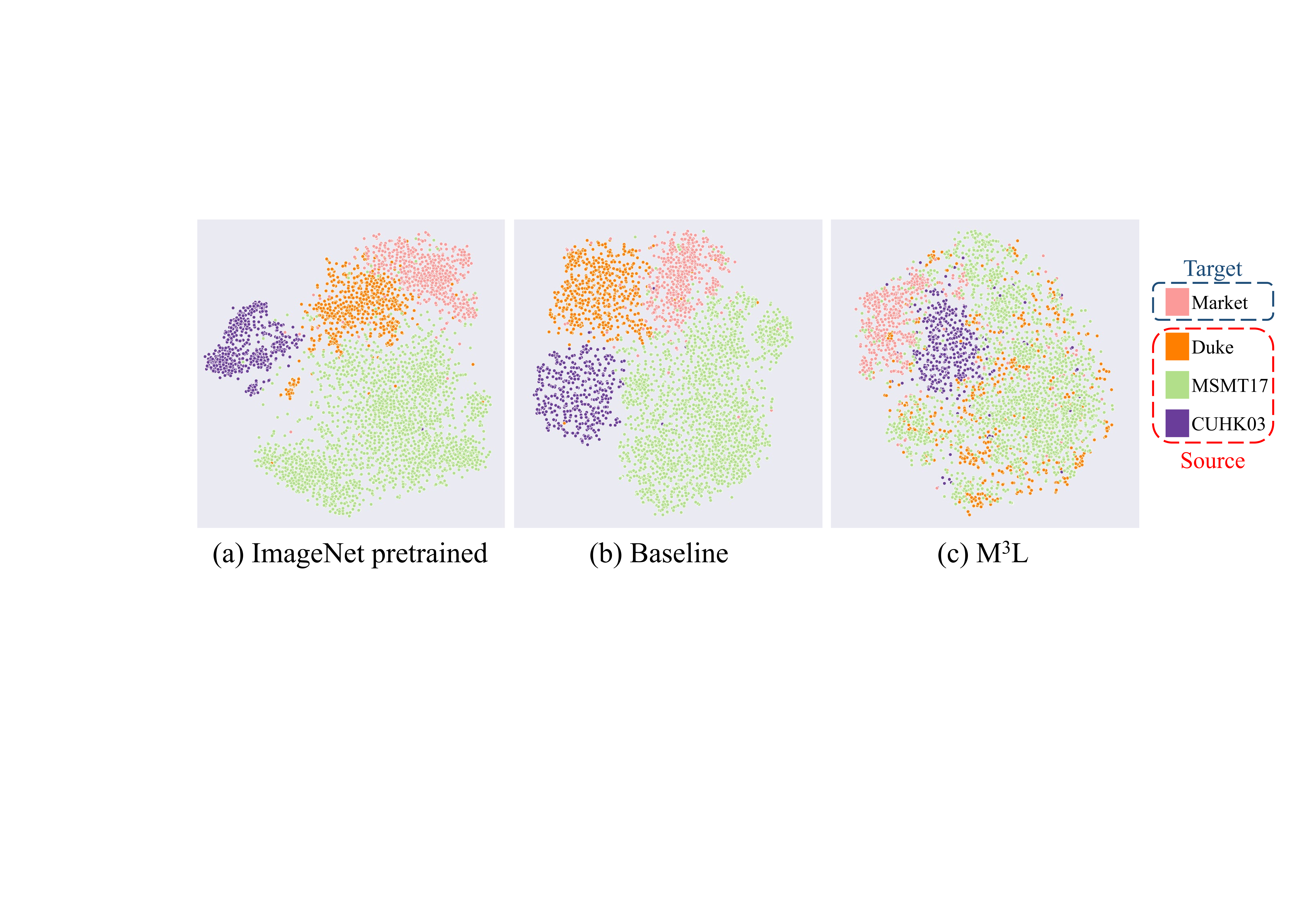}
		\vspace{-.2in}
		\caption{Visual distributions of four person ReID benchmarks. The distributions are obtained from inference features of (a)~ImageNet pretrained model, (b) Baseline, and (c) M$^3$L. All of the models are trained with ResNet-50, and the dimension of inference features is reduced by t-SNE~\cite{tsne}.}
		\vspace{-.02in}
		\label{fig:distribution}
	\end{figure}
To better understand the effectiveness of our approach, we visualize the t-SNE~\cite{tsne} distributions of the features on the four datasets for different models, \textit{i.e.}, ImageNet pretrained model, baseline, and M$^3$L. Results are shown in Fig.~\ref{fig:distribution}.	
As shown in Fig.~\ref{fig:distribution}(a), without training, distributions vary by domains: (1) MSMT17 is the largest dataset that contains images in a variety of situations; (2) DukeMTMC-reID and Market-1501 are closely related to MSMT17 and each other; (3) CUHK03 has a relatively more distinct distribution compared with the others. 
Fig.~\ref{fig:distribution}(b) and Fig.~\ref{fig:distribution}(c) show the visual distributions of the four datasets after training. The model is trained on DukeMTMC-reID, CUHK03, and MSMT17, and tested on Market-1501.
Comparing Fig.~\ref{fig:distribution}(b) with Fig.~\ref{fig:distribution}(c), we observe that the features from the source and target domains of M$^3$L~(Fig.~\ref{fig:distribution}(c)) are clustered more compactly, indicating that M$^3$L leads the model to learn more generic and domain-agnostic representations.
	
\section{Conclusion}
In this paper, we propose a Memory-based Multi-source Meta-Learning (M$^3$L) framework for multi-source domain generalization (DG) in person ReID. The proposed meta-learning strategy enables the model to simulate the train-test process of DG during training, which can efficiently improve the generalization ability of the model on unseen domains. Besides, we introduce a memory-based module and MetaBN to take full advantage of meta-learning and obtain further improvement. 
Extensive experiments demonstrate the effectiveness of our framework for training a generalizable ReID model. Our method achieves state-of-the-art generalization results on four large-scale benchmarks.

\vspace*{\baselineskip}

{\noindent\textbf{Acknowledgements}} This work is supported by the National Nature Science Foundation of China (No. 61876159, 61806172, 61662024, 62076116 and U1705286); the China Postdoctoral Science Foundation Grant (No. 2019M652257); the Fundamental Research Funds for the Central Universities (Xiamen University, No. 20720200030); the EU H2020 SPRING No. 871245 and AI4Media No. 951911 projects; and the Italy-China collaboration project TALENT: 2018YFE0118400.

\appendix
\renewcommand\thefigure{\Alph{section}} 
\renewcommand\thetable{\Alph{section}}
\renewcommand\thesection{\Alph{section}}
\section*{Appendix}
\section{Visualization of the target domain}
\begin{figure}[h]
		\centering
		\includegraphics[width=0.45\textwidth]{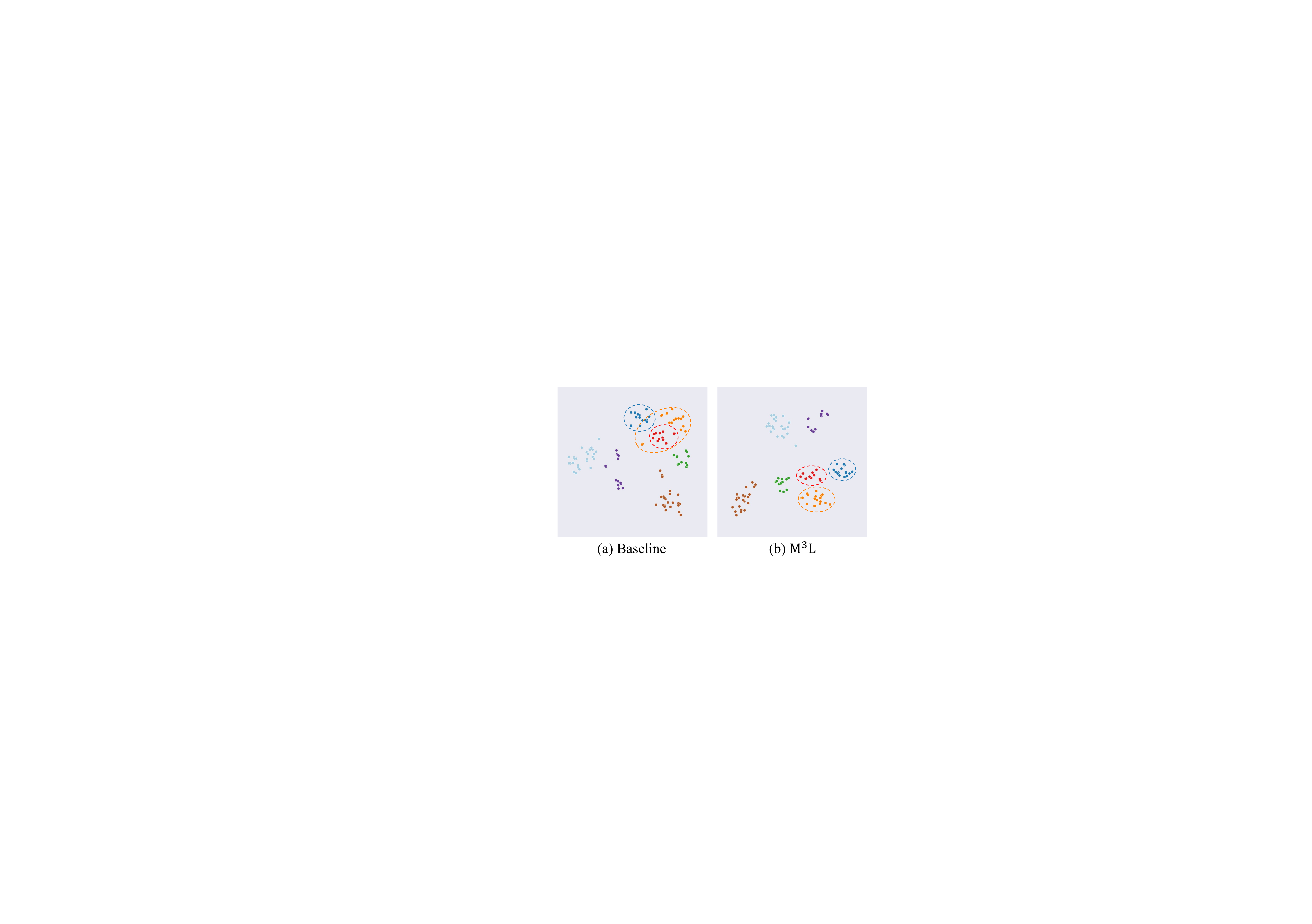}
		\vspace{-3mm}
		\caption{t-SNE~\cite{tsne} visualization of 7 persons in the unseen target dataset (Market-1501 testing set). The color indicates the identity. Results are evaluated on (a) baseline and (b) M$^3$L, both of which are trained with ResNet-50.}

		\label{fig:tsne}
	\end{figure}

To better understand the advantage of the meta-learning strategy, we visualize the distributions of the inference features of the target domain~(Market-1501 testing set) in Fig.~\ref{fig:tsne}.
Both baseline and M$^3$L are trained with DukeMTMC-reID, CUHK03, and MSMT17, and
the inference features are obtained by 7 persons in the Market-1501 testing set. We use t-SNE~\cite{tsne} to reduce the features into a 2-D space. Different colors denote different identities.
As shown in Fig.~\ref{fig:tsne}, compared with the baseline, our M$^3$L pushes the features of the same identity more compact and pull the features of different identities more discriminating.
This suggests that the proposed M$^3$L leads the model to learn more generalizable representations that can perform well on unseen domains.

\begin{table}[ht]
\centering
\caption{Results on different number of training IDs (the improvement in \textcolor{red}{red}).}
\label{tab:id}
\vspace{-3mm}
\fontsize{8pt}{9pt}\selectfont
\begin{tabular}{p{0.65cm}|p{0.45cm}<{\centering}|p{0.75cm}<{\centering}p{0.75cm}<{\centering}p{0.75cm}<{\centering}p{0.75cm}<{\centering}p{0.75cm}<{\centering}}
     \toprule
		\multirow{2}{*}{Loss} & \multirow{2}{*}{Meta} & \multicolumn{5}{c}{$\#$training IDs ~~(D+MS+C$\rightarrow$M) ~~ Rank-1 accuracy} \\
		& & 50 & 100  & 500 & 1,000 & 3,110\\
		\midrule
        \multirow{2}{*}{$\mathcal{L}_{FCG}$} & $\times$ & 27.1 & 35.2 & 48.5 &54.5 & 67.0 \\
        & \checkmark & \textbf{30.2}~(\textbf{\textcolor{red}{3.1}}) & \textbf{37.3}~(\textbf{\textcolor{red}{2.1}}) &  49.0~(\textcolor{red}{0.5}) & 55.3~(\textcolor{red}{0.8}) & 68.3~(\textcolor{red}{1.3})\\
        \midrule
        \multirow{2}{*}{$\mathcal{L}_{FCP}$} & $\times$ & 27.1 & 35.2 & 48.5 &54.5 & 67.0\\
        & \checkmark & 28.7~(\textcolor{red}{1.6}) & 36.8~(\textcolor{red}{1.6}) &  49.4~(\textcolor{red}{0.9}) & 55.9~(\textcolor{red}{1.4}) & 69.3~(\textcolor{red}{2.3})\\
        \midrule
        \multirow{2}{*}{$\mathcal{L}_{M}$} & $\times$ & 27.6 & 34.6  & 51.6 &59.6 & 67.9 \\
        & \checkmark & 28.0~(\textcolor{red}{0.4}) & 35.3~(\textcolor{red}{0.7}) & \textbf{53.8}~(\textbf{\textcolor{red}{2.2}}) &\textbf{63.4}~(\textbf{\textcolor{red}{3.8}}) & \textbf{74.5}~(\textbf{\textcolor{red}{6.6}})\\
        \bottomrule
\end{tabular}
\vspace{-.1in}
\end{table}
\section{Detailed comparison of different classifiers}
Meta-learning is effective with the FC-based classifiers in many tasks, \textit{e.g.,} few-shot learning~\cite{finn2017model,Li2018MLDG}. However, we found that the advantage of meta-learning with the FC-based classifiers will be degraded when the number of classes~(IDs) is large. In Table~\ref{tab:id}, we compare the results of three kinds of classifiers with different number of training IDs~($\#$training IDs).  
With fewer IDs, the FC-based classifiers achieve higher improvement. However, with the increase of IDs, the memory-based classifier gains higher improvement. Hence, we conclude that the FC-based classifiers are not suitable for meta-learning when the the number of classes is large. Thus, the large number of IDs in ReID leads the FC-based classifiers to produce inferior improvements than the memory-based classifier with meta-learning.

	{\small
		\bibliographystyle{ieee_fullname}
		\bibliography{reference}

\begin{thebibliography}{10}\itemsep=-1pt

\bibitem{andrychowicz2016learning}
Marcin Andrychowicz, Misha Denil, Sergio Gomez, Matthew~W Hoffman, David Pfau,
  Tom Schaul, Brendan Shillingford, and Nando De~Freitas.
\newblock Learning to learn by gradient descent by gradient descent.
\newblock In {\em NeurIPS}, 2016.

\bibitem{balaji2018metareg}
Yogesh Balaji, Swami Sankaranarayanan, and Rama Chellappa.
\newblock Metareg: Towards domain generalization using meta-regularization.
\newblock In {\em NeurIPS}, 2018.

\bibitem{2020EccvDMG}
Prithvijit Chattopadhyay, Yogesh Balaji, and Judy Hoffman.
\newblock Learning to balance specificity and invariance for in and out of
  domain generalization.
\newblock In {\em ECCV}, 2020.

\bibitem{chen2019abd}
Tianlong Chen, Shaojin Ding, Jingyi Xie, Ye Yuan, Wuyang Chen, Yang Yang, Zhou
  Ren, and Zhangyang Wang.
\newblock Abd-net: Attentive but diverse person re-identification.
\newblock In {\em CVPR}, 2019.

\bibitem{chen2019instance}
Yanbei Chen, Xiatian Zhu, and Shaogang Gong.
\newblock Instance-guided context rendering for cross-domain person
  re-identification.
\newblock In {\em ICCV}, 2019.

\bibitem{fan2018unsupervised}
Hehe Fan, Liang Zheng, Chenggang Yan, and Yi Yang.
\newblock Unsupervised person re-identification: Clustering and fine-tuning.
\newblock {\em TOMM}, 2018.

\bibitem{finn2017model}
Chelsea Finn, Pieter Abbeel, and Sergey Levine.
\newblock Model-agnostic meta-learning for fast adaptation of deep networks.
\newblock In {\em ICML}, 2017.

\bibitem{fu2019self}
Yang Fu, Yunchao Wei, Guanshuo Wang, Yuqian Zhou, Honghui Shi, and Thomas~S
  Huang.
\newblock Self-similarity grouping: A simple unsupervised cross domain
  adaptation approach for person re-identification.
\newblock In {\em ICCV}, 2019.

\bibitem{ge2020self}
Yixiao Ge, Dapeng Chen, Feng Zhu, Rui Zhao, and Hongsheng Li.
\newblock Self-paced contrastive learning with hybrid memory for domain
  adaptive object re-id.
\newblock In {\em NeurIPS}, 2020.

\bibitem{guo2020learning}
Jianzhu Guo, Xiangyu Zhu, Chenxu Zhao, Dong Cao, Zhen Lei, and Stan~Z Li.
\newblock Learning meta face recognition in unseen domains.
\newblock In {\em CVPR}, 2020.

\bibitem{resnet}
Kaiming He, Xiangyu Zhang, Shaoqing Ren, and Jian Sun.
\newblock Deep residual learning for image recognition.
\newblock In {\em CVPR}, 2016.

\bibitem{triplet}
Alexander Hermans, Lucas Beyer, and Bastian Leibe.
\newblock In defense of the triplet loss for person re-identification.
\newblock {\em arXiv:1703.07737}, 2017.

\bibitem{bn}
Sergey Ioffe and Christian Szegedy.
\newblock Batch normalization: Accelerating deep network training by reducing
  internal covariate shift.
\newblock In {\em ICML}, 2015.

\bibitem{jin2020style}
Xin Jin, Cuiling Lan, Wenjun Zeng, Zhibo Chen, and Li Zhang.
\newblock Style normalization and restitution for generalizable person
  re-identification.
\newblock In {\em CVPR}, 2020.

\bibitem{khosla2012undoing}
Aditya Khosla, Tinghui Zhou, Tomasz Malisiewicz, Alexei~A Efros, and Antonio
  Torralba.
\newblock Undoing the damage of dataset bias.
\newblock In {\em ECCV}, 2012.

\bibitem{kumar2019fairest}
Devinder Kumar, Parthipan Siva, Paul Marchwica, and Alexander Wong.
\newblock Fairest of them all: Establishing a strong baseline for cross-domain
  person reid.
\newblock {\em arXiv:1907.12016}, 2019.

\bibitem{Li2018MLDG}
Da Li, Yongxin Yang, Yi-Zhe Song, and Timothy Hospedales.
\newblock Learning to generalize: Meta-learning for domain generalization.
\newblock In {\em AAAI}, 2018.

\bibitem{Li_2019_ICCV}
Da Li, Jianshu Zhang, Yongxin Yang, Cong Liu, Yi-Zhe Song, and Timothy~M.
  Hospedales.
\newblock Episodic training for domain generalization.
\newblock In {\em ICCV}, 2019.

\bibitem{li2016learning}
Ke Li and Jitendra Malik.
\newblock Learning to optimize.
\newblock In {\em ICLR}, 2017.

\bibitem{cuhk03}
Wei Li, Rui Zhao, Tong Xiao, and Xiaogang Wang.
\newblock Deepreid: Deep filter pairing neural network for person
  re-identification.
\newblock In {\em CVPR}, 2014.

\bibitem{li2019feature}
Yiying Li, Yongxin Yang, Wei Zhou, and Timothy~M Hospedales.
\newblock Feature-critic networks for heterogeneous domain generalization.
\newblock In {\em ICML}, 2019.

\bibitem{Liao2020QAConv}
Shengcai Liao and Ling Shao.
\newblock Interpretable and generalizable person re-identification with
  query-adaptive convolution and temporal lifting.
\newblock In {\em ECCV}, 2020.

\bibitem{lin2019bottom}
Yutian Lin, Xuanyi Dong, Liang Zheng, Yan Yan, and Yi Yang.
\newblock A bottom-up clustering approach to unsupervised person
  re-identification.
\newblock In {\em AAAI}, 2019.

\bibitem{lin2020unsupervised}
Yutian Lin, Lingxi Xie, Yu Wu, Chenggang Yan, and Qi Tian.
\newblock Unsupervised person re-identification via softened similarity
  learning.
\newblock In {\em CVPR}, 2020.

\bibitem{tsne}
Laurens van~der Maaten and Geoffrey Hinton.
\newblock Visualizing data using t-sne.
\newblock {\em JMLR}, 2008.

\bibitem{muandet2013domain}
Krikamol Muandet, David Balduzzi, and Bernhard Sch{\"o}lkopf.
\newblock Domain generalization via invariant feature representation.
\newblock In {\em ICML}, 2013.

\bibitem{nichol2018first}
Alex Nichol, Joshua Achiam, and John Schulman.
\newblock On first-order meta-learning algorithms.
\newblock {\em arXiv:1803.02999}, 2018.

\bibitem{ibn}
Xingang Pan, Ping Luo, Jianping Shi, and Xiaoou Tang.
\newblock Two at once: Enhancing learning and generalization capacities via
  ibn-net.
\newblock In {\em ECCV}, 2018.

\bibitem{qiao2020learning}
Fengchun Qiao, Long Zhao, and Xi Peng.
\newblock Learning to learn single domain generalization.
\newblock In {\em CVPR}, 2020.

\bibitem{dukemtmc2}
Ergys Ristani, Francesco Solera, Roger Zou, Rita Cucchiara, and Carlo Tomasi.
\newblock Performance measures and a data set for multi-target, multi-camera
  tracking.
\newblock In {\em ECCVW}, 2016.

\bibitem{han2018coteaching}
Jun Shu, Qi Xie, Lixuan Yi, Qian Zhao, Sanping Zhou, Zongben Xu, and Deyu Meng.
\newblock Meta-weight-net: Learning an explicit mapping for sample weighting.
\newblock In {\em NeurIPS}, 2019.

\bibitem{snell2017prototypical}
Jake Snell, Kevin Swersky, and Richard Zemel.
\newblock Prototypical networks for few-shot learning.
\newblock In {\em NeurIPS}, 2017.

\bibitem{song2019generalizable}
Jifei Song, Yongxin Yang, Yi-Zhe Song, Tao Xiang, and Timothy~M Hospedales.
\newblock Generalizable person re-identification by domain-invariant mapping
  network.
\newblock In {\em CVPR}, 2019.

\bibitem{suh2018part}
Yumin Suh, Jingdong Wang, Siyu Tang, Tao Mei, and Kyoung Mu~Lee.
\newblock Part-aligned bilinear representations for person re-identification.
\newblock In {\em ECCV}, 2018.

\bibitem{sun2019meta}
Qianru Sun, Yaoyao Liu, Tat-Seng Chua, and Bernt Schiele.
\newblock Meta-transfer learning for few-shot learning.
\newblock In {\em CVPR}, 2019.

\bibitem{tay2019aanet}
Chiat-Pin Tay, Sharmili Roy, and Kim-Hui Yap.
\newblock Aanet: Attribute attention network for person re-identifications.
\newblock In {\em CVPR}, 2019.

\bibitem{thrun1998learning}
Sebastian Thrun and Lorien Pratt.
\newblock Learning to learn: Introduction and overview.
\newblock In {\em Learning to learn}. Springer, 1998.

\bibitem{vinyals2016matching}
Oriol Vinyals, Charles Blundell, Timothy Lillicrap, Daan Wierstra, et~al.
\newblock Matching networks for one shot learning.
\newblock In {\em NeurIPS}, 2016.

\bibitem{wang2020unsupervised}
Dongkai Wang and Shiliang Zhang.
\newblock Unsupervised person re-identification via multi-label classification.
\newblock In {\em CVPR}, 2020.

\bibitem{wang2018learning}
Guanshuo Wang, Yufeng Yuan, Xiong Chen, Jiwei Li, and Xi Zhou.
\newblock Learning discriminative features with multiple granularities for
  person re-identification.
\newblock In {\em ACM MM}, 2018.

\bibitem{msmt17}
Longhui Wei, Shiliang Zhang, Wen Gao, and Qi Tian.
\newblock Person transfer gan to bridge domain gap for person
  re-identification.
\newblock In {\em CVPR}, 2018.

\bibitem{memory}
Tong Xiao, Shuang Li, Bochao Wang, Liang Lin, and Xiaogang Wang.
\newblock Joint detection and identification feature learning for person
  search.
\newblock In {\em CVPR}, 2017.

\bibitem{zhai2020ad}
Yunpeng Zhai, Shijian Lu, Qixiang Ye, Xuebo Shan, Jie Chen, Rongrong Ji, and
  Yonghong Tian.
\newblock Ad-cluster: Augmented discriminative clustering for domain adaptive
  person re-identification.
\newblock In {\em CVPR}, 2020.

\bibitem{zhai2020multiple}
Yunpeng Zhai, Qixiang Ye, Shijian Lu, Mengxi Jia, Rongrong Ji, and Yonghong
  Tian.
\newblock Multiple expert brainstorming for domain adaptive person
  re-identification.
\newblock In {\em ECCV}, 2020.

\bibitem{zhang2020relation}
Zhizheng Zhang, Cuiling Lan, Wenjun Zeng, Xin Jin, and Zhibo Chen.
\newblock Relation-aware global attention for person re-identification.
\newblock In {\em CVPR}, 2020.

\bibitem{market1501}
Liang Zheng, Liyue Shen, Lu Tian, Shengjin Wang, Jingdong Wang, and Qi Tian.
\newblock Scalable person re-identification: A benchmark.
\newblock In {\em CVPR}, 2015.

\bibitem{zheng2019joint}
Zhedong Zheng, Xiaodong Yang, Zhiding Yu, Liang Zheng, Yi Yang, and Jan Kautz.
\newblock Joint discriminative and generative learning for person
  re-identification.
\newblock In {\em CVPR}, 2019.

\bibitem{dukemtmc}
Zhedong Zheng, Liang Zheng, and Yi Yang.
\newblock Unlabeled samples generated by gan improve the person
  re-identification baseline in vitro.
\newblock In {\em ICCV}, 2017.

\bibitem{rerank}
Zhun Zhong, Liang Zheng, Donglin Cao, and Shaozi Li.
\newblock Re-ranking person re-identification with k-reciprocal encoding.
\newblock In {\em CVPR}, 2017.

\bibitem{zhong2020random}
Zhun Zhong, Liang Zheng, Guoliang Kang, Shaozi Li, and Yi Yang.
\newblock Random erasing data augmentation.
\newblock In {\em AAAI}, 2020.

\bibitem{zhong2018generalizing}
Zhun Zhong, Liang Zheng, Shaozi Li, and Yi Yang.
\newblock Generalizing a person retrieval model hetero-and homogeneously.
\newblock In {\em ECCV}, 2018.

\bibitem{zhong2019invariance}
Zhun Zhong, Liang Zheng, Zhiming Luo, Shaozi Li, and Yi Yang.
\newblock Invariance matters: Exemplar memory for domain adaptive person
  re-identification.
\newblock In {\em CVPR}, 2019.

\bibitem{zhong2020memory}
Zhun Zhong, Liang Zheng, Zhiming Luo, Shaozi Li, and Yi Yang.
\newblock Learning to adapt invariance in memory for person re-identification.
\newblock {\em TPAMI}, 2020.

\bibitem{zhong2019camstyle}
Zhun Zhong, Liang Zheng, Zhedong Zheng, Shaozi Li, and Yi Yang.
\newblock Camstyle: A novel data augmentation method for person
  re-identification.
\newblock {\em TIP}, 2019.

\bibitem{zhou2019omni}
Kaiyang Zhou, Yongxin Yang, Andrea Cavallaro, and Tao Xiang.
\newblock Omni-scale feature learning for person re-identification.
\newblock In {\em ICCV}, 2019.

\bibitem{zhou2019learning}
Kaiyang Zhou, Xiatian Zhu, Yongxin Yang, Andrea Cavallaro, and Tao Xiang.
\newblock Learning generalisable omni-scale representations for person
  re-identification.
\newblock {\em arXiv:1910.06827}, 2019.

\bibitem{zou2020joint}
Yang Zou, Xiaodong Yang, Zhiding Yu, BVK Kumar, and Jan Kautz.
\newblock Joint disentangling and adaptation for cross-domain person
  re-identification.
\newblock In {\em ECCV}, 2020.

\end{thebibliography}
	}
\end{document}